\documentclass[conference, letterpaper]{IEEEtran}
\IEEEoverridecommandlockouts
\usepackage{amsmath,amssymb,amsfonts}
\usepackage{algorithmic}
\usepackage{graphicx}
\usepackage{textcomp}
\usepackage{xcolor}
\usepackage{booktabs}
\usepackage[doi=false,isbn=false,url=false,date=year,backend=biber,style=ieee,sorting=none,maxnames=3,minnames=1]{biblatex} 
\usepackage{anyfontsize}
\usepackage{enumitem}
\usepackage{mathtools}
\usepackage{dblfloatfix} 
\AtBeginBibliography{\small}
\newcommand{\required}{r}
\newcommand{\optimal}{o}
\newcommand{\metrics}{\mathcal{M}}
\newcommand{\scenarios}{\mathcal{S}}
\newcommand{\horizons}{\mathcal{H}}

\newcommand{\optimalHorizon}{\optimal_m^s}

\DeclareSourcemap{ 
    \maps[datatype=bibtex]{
      \map{
      \step[fieldsource = booktitle,
          match = Computer Vision and Pattern Recognition,
          replace = CVPR]
      \step[fieldsource = booktitle,
          match = International Conference on Computer Vision,
          replace = ICCV]
      \step[fieldsource = journal,
          match = Intelligent Transportation Systems,
          replace = ITS]
      \step[fieldsource = booktitle,
          match = Intelligent Vehicles Symposium,
          replace = IV Symposium]
      \step[fieldsource = journal,
          match = Intelligent Vehicles Symposium,
          replace = IV Symposium]
      \step[fieldsource = booktitle,
          match = International Conference on Intelligent Transportation Systems,
          replace = ITSC]
      \step[fieldsource = journal,
          match = International Conference on Intelligent Transportation Systems,
          replace = ITSC]
        }
    }      
}

\def\BibTeX{{\rm B\kern-.05em{\sc i\kern-.025em b}\kern-.08em
    T\kern-.1667em\lower.7ex\hbox{E}\kern-.125emX}}
\begin{document}


\title{Prediction Horizon Requirements for Automated Driving: Optimizing Safety, Comfort, and Efficiency \\
\thanks{This work was supported by SAFE-UP under EU’s Horizon 2020 research
and innovation programme, grant agreement 861570.}
}

\author{\IEEEauthorblockN{Manuel Muñoz Sánchez\IEEEauthorrefmark{1}, Chris van der Ploeg\IEEEauthorrefmark{4}, Robin Smit\IEEEauthorrefmark{4}, Jos Elfring\IEEEauthorrefmark{1}\IEEEauthorrefmark{2},\\Emilia Silvas\IEEEauthorrefmark{1}\IEEEauthorrefmark{4}, and René van de Molengraft\IEEEauthorrefmark{1}}
\IEEEauthorblockA{
\textit{Department of Mechanical Engineering, \IEEEauthorrefmark{1}Robotics Group, \IEEEauthorrefmark{4}Control Systems Technology Group,} \\
\textit{Eindhoven University of Technology, Eindhoven, The Netherlands}
}
\IEEEauthorblockA{\IEEEauthorrefmark{4}
\textit{Department of Integrated Vehicle Safety, TNO, Helmond, The Netherlands} \\
}
\IEEEauthorblockA{\IEEEauthorrefmark{2}
\textit{VDL CropTeq Robotics, Eindhoven, The Netherlands}
}
\normalsize Email: \texttt{m.munoz.sanchez@tue.nl}
}




\maketitle

\begin{abstract}
Predicting the movement of other road users is beneficial for improving automated vehicle (AV) performance.  
However, the relationship between the time horizon associated with these predictions and AV performance remains unclear. Despite the existence of numerous trajectory prediction algorithms, no studies have been conducted on how varying prediction lengths affect AV safety and other vehicle performance metrics, resulting in undefined horizon requirements for prediction methods. 
Our study addresses this gap by examining the effects of different prediction horizons on AV performance, focusing on safety, comfort, and efficiency. Through multiple experiments using a state-of-the-art, risk-based predictive trajectory planner, we simulated predictions with horizons up to 20 seconds. 
Based on our simulations, we propose a framework for specifying the minimum required and optimal prediction horizons based on specific AV performance criteria and application needs. 
Our results indicate that a horizon of 1.6 seconds is required to prevent collisions with crossing pedestrians, horizons of 7-8 seconds yield the best efficiency, and horizons up to 15 seconds improve passenger comfort. 
We conclude that prediction horizon requirements are application-dependent, and recommend aiming for a prediction horizon of 11.8 seconds as a general guideline for applications involving crossing pedestrians.



\end{abstract}


\section{Introduction}
\begin{figure}[t]
    \centering
    \includegraphics[width=\linewidth]{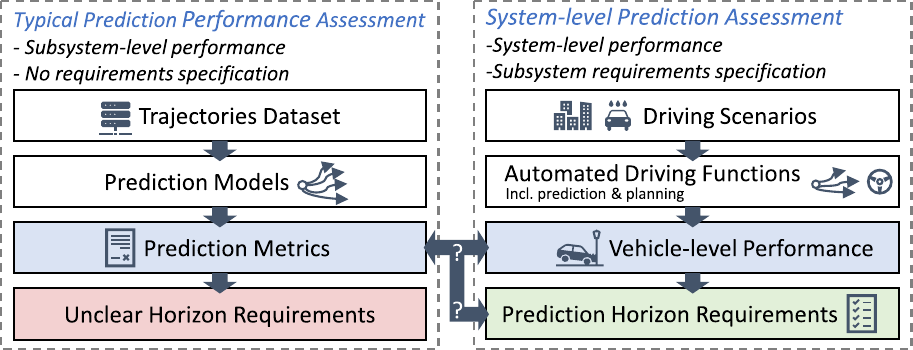}
    \vspace{-18pt}
    \caption{Typical assessment of trajectory prediction work (left) and our approach (right).}\label{fig:front}%
    \vspace{-16pt}
\end{figure}

Automated vehicles (AVs) are becoming increasingly prevalent, and trajectory prediction of surrounding road users (RUs) is a critical component of these systems, since the AV's decisions will be largely based on the predicted motion of other RUs. 
%
When designing a system that accounts for the future motion of surrounding RUs, one must decide how much into the future to predict, i.e. the prediction horizon, which is used to plan an AV trajectory for the same horizon. The choice of prediction horizon can have a significant impact on AV behavior, however, in the development of prediction models the chosen horizon is often an ad hoc decision that depends on practical aspects, such as the length of traces available in the data used to develop these models. 
Additionally, the typical assessment of prediction models is performed at the subsystem-level, i.e. the prediction module, focusing on predictive accuracy~\cite{Rudenko2020HumanSurvey}. However, with this approach, the impact of predictions on the overall system, namely the AV, remains unclear (Fig~\ref{fig:front}). Consequently, prediction horizon requirements remain undefined. 

Despite the existence of several works showing that it is beneficial to integrate predictions in motion planning~\cite{Hagedorn2023RethinkingReview}, there is no clear quantification of the relationship between the prediction horizon and the resulting impact on AV behavior. To establish requirements that a prediction model should adhere to, 
the impact that predictions have on AV behavior must first be understood. 
Once we understand this impact, we can determine the ideal prediction horizons that ensure safety and other key aspects like ride comfort and efficiency. 

The requirements that predictions must satisfy will be dependent on several other factors, like the type of object being predicted (e.g. vehicle vs. pedestrian), the specific scenario (e.g. lane change in a highway vs a crossing pedestrian), or even user preferences (e.g. a smoother ride is preferred over lower travel time). Hence, these requirements are application-dependent and should adapt to each situation.

To specify prediction requirements, this work first investigates the impact that different prediction horizons have on the safety, comfort and efficiency of an AV. We do this by simulating and integrating predictions into a state of the art optimization-based planner that considers the predicted actions of other RUs~\cite{vanderPloeg2023ConnectingReasoning}. Additionally, we introduce a framework to establish both the minimum required and optimal prediction horizons for achieving targeted vehicle performance in specific applications. To show the applicability of our method, we focus on urban driving scenarios involving crossing pedestrians, which are among the most vulnerable participants in traffic~\cite{D2.6}.




Our main contributions are summarized as follows:

\begin{itemize}[leftmargin=*]
\item \textit{Impact analysis of trajectory prediction horizons}: We explore the relationship between different prediction horizons and the safety, comfort, and efficiency of AVs in scenarios involving crossing pedestrians.


\item \textit{A methodology to derive application-specific requirements}: We introduce a versatile framework to determine prediction horizon requirements tailored to specific AV applications and performance goals. While we demonstrate its application in
a limited set of scenarios and performance criteria, 
the framework is adaptable to other scenarios and criteria.


\item \textit{Guidelines for required and optimal prediction horizons}: Using our framework, we provide recommendations for the ideal horizon of trajectory prediction models in AV applications involving crossing pedestrians.

\end{itemize}

The remainder of this article is structured as follows. Section~\ref{sec:related-work} highlights prevalent choices of prediction horizons, related work combining predictions and motion planning, and vehicle-level metrics commonly used to evaluate AV performance. Section~\ref{sec:methodology} introduces our proposed methodology to derive horizon requirements. Section~\ref{sec:results} summarizes the results and
highlights limitations of this study. Finally, Section~\ref{sec:conclusion} concludes the work and outlines future improvements.




\section{Related Work}\label{sec:related-work}
This section offers a review of prevalent choices of prediction horizon. Subsequently, we analyze works considering predictions and motion planning jointly. Finally, we delve into the metrics that can be used to evaluate AV performance.

\subsection{Choosing a Prediction Horizon}
The field of trajectory prediction is dominated by machine learning methods nowadays~\cite{Rudenko2020HumanSurvey}. The development of these methods relies on datasets with pre-recorded RU trajectories, segmented in two parts with an observation and a prediction horizon, i.e.  the portion of the trajectory available before prediction, and the portion to be predicted, respectively.

Table~\ref{tab:horizons} presents an overview of popular datasets commonly used for trajectory prediction research. Methods using these datasets seem to be conditioned on either \textit{(i)} the length of the traces contained in the dataset, or \textit{(ii)} popular choice of horizons for a specific dataset used in previous work, which facilitates comparison of a new method without the need to re-implement previous ones.
\begin{table}[htb]
\centering
\caption{Popular trajectory prediction datasets and associated (maximum) prediction horizons reported \cite{Ivanovic2023Trajdata:Datasets,Rudenko2020HumanSurvey,Moers2022TheGermany}}
\label{tab:horizons} \vspace{-6pt}
\begin{tabular}{@{}rccl@{}} 
\toprule
Dataset   & Year & Horizon {[}s{]}  \\ \midrule
UCY         &   2007   &  4.8             \\ 
NGSIM       &   2007   &          5               \\
ETH         &  2009    &      4.8                   \\
SDD         &   2016   &         4.8                \\
highD      &   2018   &     5                  \\
ApolloScape Trajectories        &  2019    &      3              \\
NuScenes    &  2019    & 6                       \\
Argoverse   &   2019   & 3                       \\
Interaction       & 2019 & 3                      \\
Lyft       &  2020    & 5                       \\
WOMD              & 2021 & 8                       \\
Shifts       &   2021   & 5                       \\
Argoverse 2   &   2021   &    6                     \\
exiD       &  2022    &     5                 \\ \bottomrule
\end{tabular}
\vspace{-16pt}
\end{table}

Since the choice of prediction horizon is arbitrary and independent from AV performance metrics, predictive accuracy is often analyzed at various horizons to obtain a better coverage of a model's capabilities. For instance, methods working with UCY, ETH and SSD typically report performance at a horizon of 4.8 seconds~\cite{Alahi2016SocialSpaces,Zhang2019SR-LSTM:Prediction,Sadeghian2019SoPhie:Constraints}, but some works also report their performance at 3.2 seconds~\cite{Gupta2018SocialNetworks}. Methods evaluated on NGSIM, highD, and exiD report from 1 up to 5 seconds~\cite{Messaoud2019Non-localPrediction,Messaoud2021AttentionPrediction,Mozaffari2023TrajectoryScenarios}. Reporting the performance at different horizons presents a more complete assessment of a model's predictive capabilities, but it remains unclear how its predictions ultimately affect AV behavior and what the required and optimal horizons are.

\subsection{Evaluating Predictions \& their Impact on Vehicle Behavior}

Predictions are mainly evaluated in terms of geometric accuracy with metrics such as minimum average displacement error, and minimum final displacement error~\cite{MunozSanchez2022}, often disregarding their associated uncertainty and the impact they have on the behavior of the vehicle. 

The authors of~\cite{Tran2023WhatDriving} take a critical look at current evaluation practices and conclude that trajectory prediction metrics do not reflect well the impact they have on AV behavior. To address this limitation, the authors of~\cite{Ivanovic2022InjectingEvaluation} advocate for a novel planning-aware metric which better reflects performance in AVs, weighting the error of each prediction according to the impact it would have on the planned AV trajectory.

Other than lacking evaluation practices, the authors of~\cite{IvanovicMATS:Control} raise concern regarding the currently accepted output representation of predictions for downstream integration, since a majority of planning and control algorithms reason about system dynamics instead of the current prediction representation, i.e. a sequence of positions. Accordingly, they propose a dynamical system representation that allows the planning optimizer to simultaneously explore different AV controls and their effect on the predictions of surrounding RU's. 
Although not as efficiently as proposed in~\cite{IvanovicMATS:Control}, the current representation allows incorporating predictions in motion planning~\cite{GeisslingerAnVehicles}, and conversely use the AV's planned trajectory to influence predictions of surrounding RUs~\cite{Dong2022Graph-basedDriving}.

Despite the existence of numerous studies coupling predictions with planning for improved AV performance, there is a noticeable gap in research quantifying these improvements~\cite{Hagedorn2023RethinkingReview}. 
A significant aspect of uncertainty involves determining the necessary and ideal prediction horizons. That is, the shortest horizon that ensures acceptable AV performance, and the horizon that optimizes AV performance. Short horizons might not allow enough time to react accordingly, and the full benefits of predictions might not be realized. Conversely, excessively long horizons could lead to overly cautious behavior or increased computational demands, potentially diminishing AV performance.

\subsection{Vehicle-level Metrics}
To assess the impact of predictions on AV behavior, vehicle-level performance metrics are needed. AV performance is often measured in terms of safety, comfort, and efficiency \cite{Sohrabi2021QuantifyingResearch,deWinkel2023StandardsJerk,Taiebat2018AVehicles}.

\subsubsection{Safety} The potential to increase safety is one of the main reasons AVs are a desirable technology. According to the National Highway Traffic Safety Administration~\cite{nhtsa2018critical}, human error is a contributing factor in 94\% of crashes, and it is expected that AVs will be able to prevent such crashes~\cite{Sohrabi2021QuantifyingResearch}. Consequently, one of the most popular metrics to assess safety is collision rate, often compared to human-driven vehicles~\cite{Petrovic2020TrafficDrivers}. Nonetheless, there are other relevant metrics used to assess the safety of an AV, such as disengagements (i.e. the number of times a human driver needed to take control of the vehicle)~\cite{Dixit2016AutonomousTimes}, and safety violations~\cite{Li2020AV-FUZZER:Systems}.

\subsubsection{Comfort} The comfort of AVs is commonly measured using ISO standard 2631~\cite{iso2631}, which provides approaches for categorizing comfort levels of vibration and shock motion at frequencies similar to those encountered in a vehicle.  However, a recent study investigated empirically the relationship between a vehicle's acceleration and jerk, and the discomfort experienced by passengers, concluding that the standard models in ISO 2631 may not effectively describe the comfort of motions experienced in a vehicle~\cite{deWinkel2023StandardsJerk}. At the same time, they concluded that acceleration is the main descriptor of (dis)comfort, and derived acceleration thresholds for seven different ranges of comfort, from excellent to terrible.

\subsubsection{Efficiency} Efficiency of an AV can be considered with respect to several aspects. 
A commonly used metric for measuring the efficiency of an AV is fuel efficiency~\cite{Jungblut2023Fuel-savingAnalysis}, which in turn can help reduce emissions~\cite{Ercan2022AutonomousCities}. Efficiency is also often measured in terms of travel time~\cite{Szimba2020AssessingRelation}, which can be considered a deciding factor for choosing a mode of transport~\cite{Wadud2023FullyUse}. 

\section{Methodology}\label{sec:methodology}
\begin{figure}[htb]
    \centering
    \includegraphics[width=\linewidth]{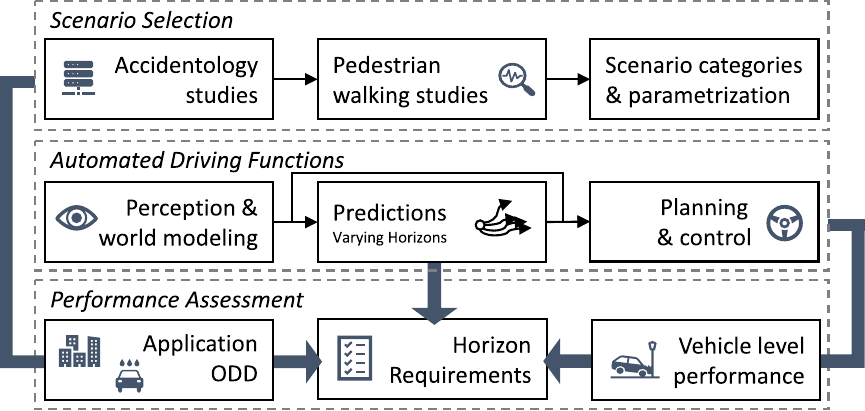}
    \vspace{-18pt}
    \caption{Methodology to derive prediction horizon requirements.}
    \label{fig:overview}
    \vspace{-12pt}
\end{figure}

This section describes our methodology to derive prediction horizon requirements. Fig.~\ref{fig:overview} presents an overview of this process. First, relevant scenarios are selected based on accidentology and pedestrian walking studies. Next, typical modules of an AV are integrated in a simulation environment considering varying prediction horizons of surrounding RUs. Finally, combining vehicle-level performance when using different horizons and the intended application operational design domain (ODD), horizon requirements are derived.

\subsection{Scenario Selection}\label{sec:sub:scenarios}
Accidentology studies highlight that vulnerable RUs such as pedestrians are among the most at-risk traffic participants in urban environments~\cite{D2.6}. Specifically, the most frequently occurring accident is a pedestrian crossing from the right without any sight obstruction. This group of accidents makes up 22.8\% of car-to-pedestrian collisions, and it is the most lethal collision, amounting to 23.2\% of all killed or severely injured pedestrians~\cite{D2.6}. In these accidents, vehicle speed ranged from 26 to 48 km/h, and pedestrian speeds are unknown. This group of accidents is selected as the basis of our study.

We investigate the impact of predictions on three similar scenario categories (SCs): \textit{SC1}, \textit{SC2}, and \textit{SC3}, where the AV speeds are set to 30, 40 and 50km/h respectively. For the three SCs, one hundred different pedestrian speeds are sampled randomly from a normal distribution describing pedestrian walking speeds, with mean 1.34m/s and standard deviation 0.37m/s~\cite{pedestrianSpeeds}. Thus, with 3 SCs, 22 different horizons, and 100 pedestrian speeds, a total of 6600 runs are executed.  

\begin{figure}[htb]
    \centering
    \includegraphics[width=\linewidth]{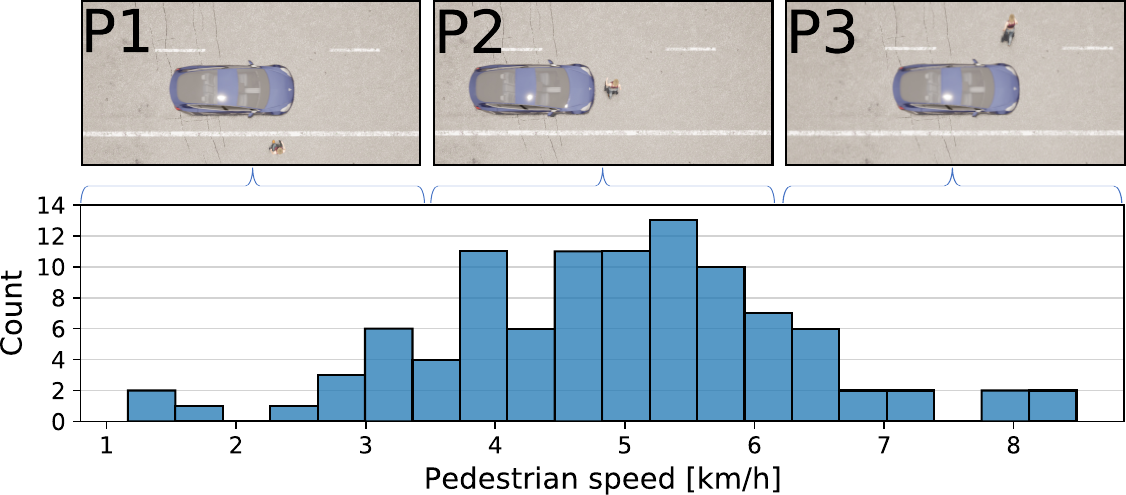}
    
    \caption{
     Histogram of pedestrian speeds used during this study and corresponding scenario when the vehicle does not react to the pedestrian.}
    \label{fig:pedestrian-speeds}
\end{figure}

The scenarios are set up such that if the AV does not react to the pedestrian, a front collision occurs in the middle of the AV when the pedestrian is walking at the mean speed of 1.34m/s. Consequently, for pedestrian speeds that significantly deviate from this value, there would be no collision even when the AV does not react to the pedestrian, due to the pedestrian walking too slow or too fast (Fig.~\ref{fig:pedestrian-speeds}). We refer to these three groups of pedestrian speeds as follows:

\begin{itemize}
\item[\textit{P1}:] Slow pedestrian speeds, such that the AV passes the pedestrian collision-free before it crosses the road.
\item[\textit{P2}:] Pedestrian speeds that result in a collision when the AV does not react to the pedestrian.
\item[\textit{P3}:] Fast pedestrian speeds, such that the AV passes the pedestrian collision-free after it has crossed the road.
\end{itemize}


\subsection{Automated Driving Functions}

\subsubsection{Perception and world modeling} 
Analyzing the impact of perception and world modeling performance on vehicle performance falls out of the scope of this study. Consequently, factors such as sensor noise and perception-related challenges are not considered. In our simulations, we gather data on nearby objects, such as road lanes and other RUs, directly from the simulator.

\subsubsection{Predictions} 

A wide range of methods exists for predicting the trajectories of RUs \cite{Rudenko2020HumanSurvey,Schuetz2023AUser}, each with distinct advantages and drawbacks. Opting for a single prediction method could skew our results in favor of that method. Additionally, trajectory prediction methods are rapidly evolving and becoming more accurate. Thus, if using currently existing prediction methods to derive their requirements, these requirements could become obsolete as the field evolves. To circumvent this, our approach involves simulating an ideal scenario, where predictions are flawless, to find the horizons that would provide the most benefits if predictions were perfect. 
Different time horizons of these ideal predictions are then provided to the planner to find the necessary and most beneficial horizons.  A thorough investigation into the effects of prediction inaccuracies and the subsequent derivation of accuracy requirements is reserved for future work.
To limit simulation times and have a good coverage of the search space, we consider horizons
\begin{equation} \label{eq:horizons}
    \horizons = \{0, 0.2, 0.4, ..., 2, 3, ..., 10, 12, 15, 20 \} \ \text{seconds,}
\end{equation}
and the achieved performance with horizons that are not in $\horizons$ is calculated by linear interpolation.

\subsubsection{Planning \& Control}
To evaluate the effect of trajectory prediction on AV behavior, a state of the art planner is integrated in our simulations~\cite{vanderPloeg2023ConnectingReasoning}. This planner considers both the road infrastructure and the predicted positions of surrounding objects. It uses artificial potential-based risk fields within the cost function of a model-predictive control problem to find a trajectory that minimizes risk while progressing towards the AV's intended destination. For a more comprehensive description of this planner, we defer the reader to~\cite{vanderPloeg2023ConnectingReasoning}.

\subsection{Performance Assessment}
In this section, we outline our methodology for AV performance assessment and derivation of prediction horizon requirements. First, we present the vehicle-level performance metrics employed in our assessment. Subsequently, the proposed framework designed to determine both required and optimal prediction horizons is introduced. Lastly, we illustrate the versatility of this framework by introducing four distinct use cases to demonstrate its application across various AV settings tailored to specific operational purposes.




\subsubsection{Vehicle-level Performance} 
\label{sec:sub:vehicle-metrics}
To assess the impact of prediction horizons on AV behavior, we measure safety, comfort and efficiency as described in this section. Additionally, we analyze the real-time capabilities of our system. 

Let $f^m(h, s)$ denote the value achieved for vehicle-level metric $m$ and SC $s$ when using a prediction horizon $h$. Each of these metrics is defined as: 

\paragraph{Safety}{
The percentage of runs that are collision-free. Since we are mainly interested in achieving collision-free runs, we do not weigh in the AV's speed at impact, but we visualize it for illustrative purposes.}

\paragraph{Comfort}{We adapt the categorization of (dis)comfort based on acceleration presented in~\cite{deWinkel2023StandardsJerk} in combination with~\cite{Bae2019} and consider deceleration values as comfortable, uncomfortable, and highly uncomfortable, as shown in Table~\ref{tab:comfort-journal}. Considering the time that the vehicle is decelerating, we report the percentage of time spent on each comfort category. The resulting comfort in SC $s$ when using a prediction horizon $h$, $f^{\text{\textit{comfort}}}(h, s)$,  is the percentage of time spent on the comfortable category.}
\begin{table}[htb]
\centering
\caption{Deceleration thresholds for (dis)comfort level~\cite{deWinkel2023StandardsJerk,Bae2019}}
\label{tab:comfort-journal}
\begin{tabular}{@{}cc@{}}
\toprule
Acceleration {[}m/s$^2${]} & Comfort Level \\ \midrule
{[}-0.89, 0)                      & Comfortable             \\
{[}-1.89, -0.89)                      & Uncomfortable      \\
\textless~-1.89                      & Highly Uncomfortable  \\ \bottomrule
\end{tabular}
\end{table}

\paragraph{Efficiency}
We investigate efficiency in terms of travel time, and calculate the relative increase of travel time, $\Delta t$, for each prediction horizon. That is, if $t_b(s)$ is the time it takes the vehicle to complete the route in scenario $s$ without a pedestrian crossing the road, and $t(h, s)$ is the time with the crossing pedestrian and using prediction horizon $h$, then
\begin{equation}
\Delta t(h,s) = 100 \frac{t(h, s) - t_b(s)}{t_b(s)}.
\end{equation}
To keep the interpretation of efficiency consistent with other metrics (i.e. within a positive range, and where a higher value of the metric denotes better performance), the final efficiency is calculated as
\begin{align}
    f^{\text{\textit{efficiency}}}(h, s) = - \Delta t(h,s) + \max_{h'\in\mathbb{R^+}} \Delta t(h',s).
\end{align}

\paragraph{Real-time execution}
The ability to operate at a high frequency is not one of the goals of this study. However, with higher prediction and planning horizons, the load on the trajectory planner impeded its operation at the desired frequency\footnote{All simulations are executed on a desktop computer with an AMD Ryzen 9 5900X processor, GeForce RTX 3070 Ti GPU, and 64GB DDR4 RAM.}, which had an impact on comfort and efficiency. Thus, we evaluate the frequency at which the planner updates its trajectories\footnote{Measured by the rate of new trajectory arrivals on the corresponding ROS topic, averaged over a 21-sample sliding window.}. For each simulation, we record the minimum frequency, and then report the mean and standard deviation of these values across the 100 runs for each horizon.

\subsubsection{Deriving Required and Optimal Prediction Horizons}  
Different scenarios and metrics can lead to significantly different horizon requirements. Thus, concluding on a preferred horizon is not trivial. This section describes a methodology to derive the preferred horizon by combining multiple system-level (i.e. vehicle) metrics in a variety of scenarios.

In our simulations, all possible vehicle-level metrics are $\metrics=\{\textit{safety}, \textit{comfort}, \textit{efficiency}\}$, all possible SCs are $\scenarios = \{\textit{SC1}, \textit{SC2}, \textit{SC3}\}$, and all possible horizons are $\horizons$ as in \eqref{eq:horizons}.
When considering a collection of metrics $M \subseteq \metrics$ and SCs $S \subseteq \scenarios$, we aim to find a \textit{required} and an \textit{optimal} horizon, $\required_M^S$, and $\optimal_M^S$, such that horizon $\required_M^S$ yields satisfactory AV performance for all metrics and scenarios considered, and horizon $\optimal_M^S$ yields the best value of each metric in every scenario. If that is not possible, we should find the horizon that yields the best trade-off, as explained later in this section.  

When considering individual metrics, our definitions of satisfactory and optimal vary depending on the metric:
\begin{itemize}[leftmargin=*]
    \item When considering only safety, i.e. $M = \{\text{\textit{safety}}\}$, $\required_M^S = \optimal_M^S $ is the shortest horizon that maximizes safety. We aim at zero collisions, and anything else is not considered satisfactory.
    \item If $M = \{\text{\textit{comfort}}\}$, $\optimal_M^S$ is the shortest horizon that maximizes the share of comfortable braking, and $\required_M^S$ is the horizon that minimizes the share of highly uncomfortable braking.
    \item If $M = \{\text{\textit{efficiency}}\}$, $\optimal_M^S$ is the shortest horizon that maximizes efficiency, and $\required_M^S$ is the shortest satisficing~\cite{Artinger2022Satisficing:Traditions} horizon, that is, the horizon achieving an efficiency that is within 15\% of the optimal efficiency\footnote{The threshold for the satisficing, i.e. near optimal ``good enough''~\cite{Artinger2022Satisficing:Traditions}, horizon is determined by visual inspection, selecting the horizon at which efficiency begins to converge. All the selected horizons yield an efficiency that is within 15\% of the optimal value. }.
\end{itemize}

Different metrics may lead to conflicting required and optimal horizons; thus, a procedure is needed to find the overall optimal and required horizons when considering multiple potentially conflicting objectives and scenarios.

\begin{figure}[htb]
    \centering
     \includegraphics[width=\linewidth]{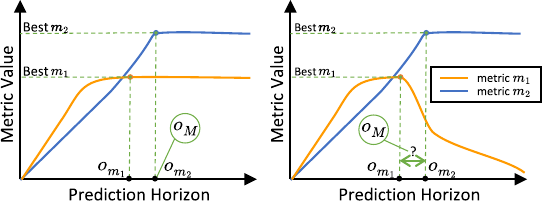}
    \caption{Example of two situations yielding a different overall optimal horizon. Left: it is possible to choose one of the optimal horizons without negatively affecting any metric. Right: it is not possible to choose an optimal horizon without negatively affecting some metric.}
    \label{fig:aggregated-horizons-optimal}
\end{figure}

\paragraph{Overall optimal horizon}
When considering multiple objectives and scenarios to conclude on the overall optimal horizon, different situations are possible. Consider as an example the two situations shown in Fig.~\ref{fig:aggregated-horizons-optimal}, where the value of two metrics $m_1$ and $m_2$ are shown as a function of the prediction horizon. 
For simplicity, different SCs are left out of the example. In the first situation (left), both metrics $m_1$ and $m_2$ converge to their optimal values at horizons $o_{m_1}$ and $o_{m_2}$. In this case, we would choose the optimal horizon, $o_M$, to be the same as $o_{m_2}$, as this is the minimum horizon with which all considered metrics achieve their optimal value. 

In the second example (Fig.~\ref{fig:aggregated-horizons-optimal}, right), however, $m_1$ is negatively affected by longer horizons, therefore choosing $o_{m_2}$ as the optimal horizon is no longer possible. Since it is not possible to choose a horizon that leads to the optimal value across all metrics and scenarios, we should find the best trade-off. The best trade-off is subjective and depends on the relative importance given to different metrics and scenarios. To find the horizon that best balances all considered metrics across all considered SCs, the cost function $f^C$ is introduced as
\begin{equation}\label{eq:aggregation:cost}
    f^C(h, S, M) =  \sum_{s\in S} \sum_{m \in M} 
 w_s w_m (\Tilde{f}^m(h, s)-\Tilde{f}^m(\optimalHorizon, s))^2\text{,}
\end{equation}
where $w_s$ and $w_m$ are weighting factors to balance the importance of each SC and metric. Recall that $f^m(h, s)$ denotes the value of metric $m$ achieved in scenario $s$ when using a prediction horizon $h$, thus, $f^m(\optimalHorizon, s)$ denotes the best value of metric $m$ in scenario $s$, achieved with horizon $\optimalHorizon$. Additionally, note that $\Tilde{f}^m$ is used instead of $f^m$, which denotes a normalized version of $f^m$ to ensure all metric values are in a common range\footnote{In our case, between 0 and 100, where 0 maps to the worst value of the metric achieved in all simulations, and 100 maps to the best.} in order to provide a more intuitive weighting scheme.

The optimal horizon over all scenario categories $S$ and metrics $M$, $\optimal_M^S$, is then calculated by obtaining the horizon that minimizes deviation from all optimal metric values for each scenario $s \in S$ and metric $m \in M$. In case the resulting horizon is not sufficient to guarantee safety, then the horizon that maximizes safety is used instead. That is,  
\begin{equation}\label{eq:agg:optimal}
\optimal_M^S = \max \left(\optimal_{\text{\textit{safety}}}^s, \underset{h\mathbb\in{R}^+}{\arg\min} f^C(h, S, M) \right). 
\end{equation}

\begin{figure}[htb]
    \centering
     \includegraphics[width=\linewidth]{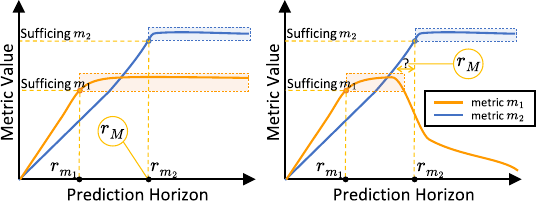}
    \caption{Example of two situations yielding a different overall required horizon. 
    Left: it is possible to choose a horizon that satisfies the required value of all metrics. Right: it is not possible to choose a horizon that satisfies the required value of all metrics.}
    
    \label{fig:aggregated-horizons-required}
\end{figure}

\paragraph{Overall required horizon} A different approach is taken to find the overall required prediction horizon. The reason for this deviation is illustrated with an example. Consider the two situations shown in Fig.~\ref{fig:aggregated-horizons-required}. 
In the first example (left), horizons beyond the required $r_{m_1}$ and  $r_{m_2}$ all yield a satisfactory value for metrics $m_1$ and $m_2$, respectively, denoted in the figure by the highlighted rectangular area. Similarly to the example from Fig.~\ref{fig:aggregated-horizons-optimal} (left), in this case the overall minimum required horizon, $r_M$, can simply be chosen the same as $r_{m_2}$. However, in Fig.~\ref{fig:aggregated-horizons-required} (right), there is no horizon that yields a satisfactory value for all metrics simultaneously. 
If this value does not exist, the best alternative is to use the horizon achieving the best trade-off across all metrics and scenarios, the optimal horizon $o_M^S$ in \eqref{eq:agg:optimal}, or the system designer should review the previously specified requirements for each vehicle-level metric. 


Thus, the overall required horizon is calculated as the shortest horizon that yields a satisfactory value for all considered metrics and scenarios, if it exists. Otherwise, no required horizon can be provided. To formalize the derivation of the required horizon, let us first introduce an auxiliary binary function $f^I$ to determine whether or not a metric $m$ in scenario $s$ is considered, based on their weights $w_s$ and $w_m$. We shall consider it if both weights are nonzero:
\begin{equation}
f^I(w_s, w_m)
    \begin{cases}
       0 \text{ if } w_s = 0 \text{ or } w_m = 0\\
        1 \text{ otherwise.}
    \end{cases}
\end{equation}
The set of horizons that yield a satisfactory value for all metrics and scenarios is given by 
\begin{equation} 
\mathcal{H}^r=\{ h \in \mathbb{R^+} \mid \forall_{s \in S, m \in M} [f^m(h, s) \geq f^m(r_m^s, s) ] \}.
\end{equation}
Then the overall required horizon, if $\mathcal{H}^r \neq \emptyset $, is given by
\begin{equation} \label{eq:req}
\required_M^S = \max \left(\optimal_{\text{\textit{safety}}}^s, \min \left(\ \mathcal{H}^r \right) \right).
\end{equation}

\begin{figure*}[htb]
    \centering
    \includegraphics[width=\textwidth]{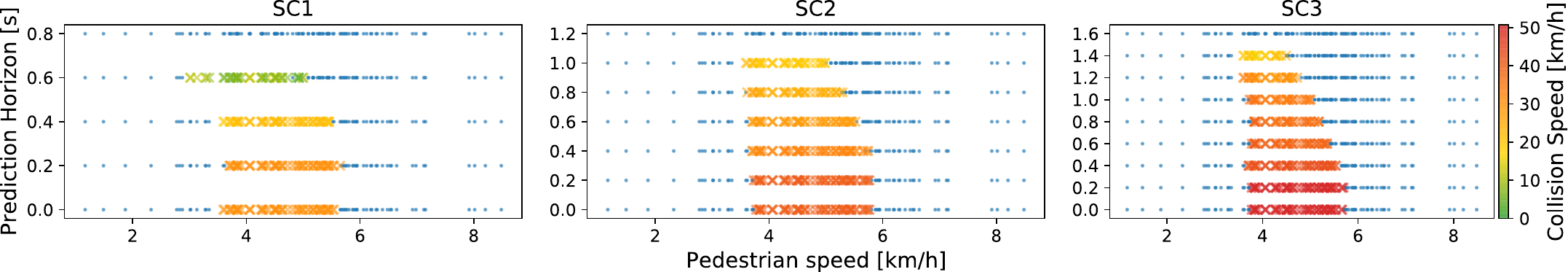}
    \vspace{-20pt}
    \caption{Collisions and the AV's speed at impact for different prediction horizons and pedestrian speeds. An ``x" denotes a collision, and its color indicates the AV speed at the moment of collision. A blue ``." denotes a run without a collision.}
    \label{fig:results-safety}
    \vspace{-10pt}
\end{figure*}
\begin{figure*}[htb]
    \centering
    \includegraphics[width=\textwidth]{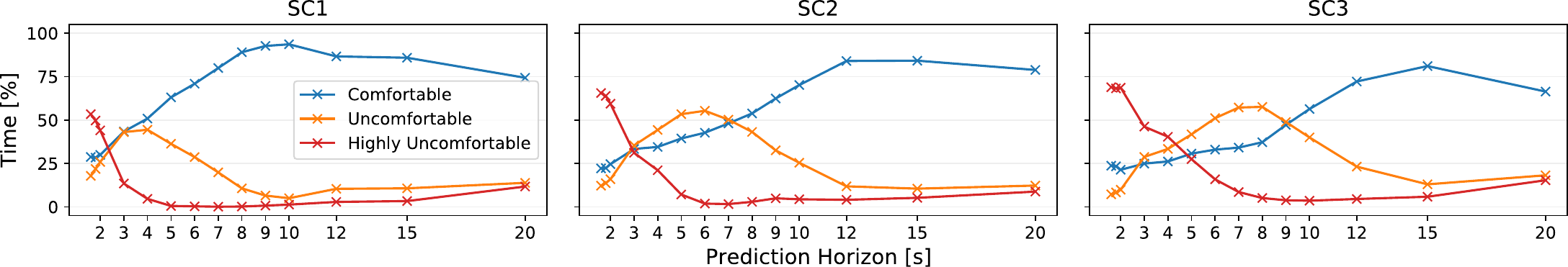}
    \vspace{-20pt}
    \caption{Percentage of time spent on different levels of comfort during braking for different horizons.}
    \label{fig:results-comfort}
    \vspace{-12pt}
\end{figure*}


\subsubsection{Application-specific requirements}
To exemplify the relevance of the intended application, and the versatility of our framework, let us consider four fictional applications:

\begin{itemize}
    \item[\textit{a)}] \textit{General-purpose urban driving}. The AV operates between 30km/h and 50km/h, and having the best trade-off between comfort and efficiency is desired. 

\item[\textit{b)}] \textit{Driverless food delivery}. 
There is no human in the AV, thus emphasis is placed on minimizing delivery time over passenger comfort.
Depending on the items being delivered, the AV is estimated to operate at 30, 40, and 50km/h 20, 10, and 70\% of the time, respectively.  

\item[\textit{c)}] \textit{Driverless taxi. Passenger is not in a hurry}. The person in the AV does not care about saving time to their destination. The AV operates between 30 and 50km/h.

\item[\textit{d)}] \textit{Driverless taxi. Passenger is in a hurry}. The person in the AV is in a hurry to get to the airport as soon as possible, even if that means sacrificing comfort. The AV operates at 50km/h whenever possible. 
\end{itemize}

With the proposed framework, it is possible to determine the necessary and ideal horizons for each specific application by customizing the weights assigned to each scenario and metric. Note that safety is not factored into this weighting, as it should not be compromised for other secondary objectives like comfort or efficiency. In the considered applications, the suggested weights might look like those outlined in Table~\ref{tab:results-weights}. 
These weights are flexible and can be modified based on user preferences or the specific application requirements. For instance, if an AV is not expected to operate at speeds of 50km/h, this can be factored in by assigning a zero weight to scenario SC3. Similarly, if a customer receiving a meal delivery (as in application \textit{b)}) values a smoother ride, potentially at the cost of longer delivery times, this preference can be reflected by adjusting the weights for comfort and efficiency accordingly. It is important to note that the values in Table~\ref{tab:results-weights} are provided merely as examples for the hypothetical applications described and are not necessarily the recommended values.
\begin{table}[htb]
\centering
\caption{Possible metric and scenario weights for each application}
\label{tab:results-weights}
\begin{tabular}{@{}c|ccccc@{}}
\toprule
 & \multicolumn{5}{c}{Weight} \\
Application & Comfort & Efficiency & SC1 & SC2 & SC3  \\ \midrule
a) & 1 & 1 & 1 & 1 & 1  \\
b) & 0 & 1 & 0.2 & 0.1 & 0.7  \\
c) & 1 & 0 & 1 & 1 & 1   \\
d) & 0.1 & 2 & 0 & 0 & 1  \\ \bottomrule
\end{tabular}
\end{table}





\section{Results}\label{sec:results}
This section first presents vehicle-level metrics with different prediction horizons. Then we analyze the real-time capabilities of our AV architecture, highlighting how extended prediction and planning horizons can influence overall AV performance negatively due to increased computational demands. Next, the proposed framework is applied to derive prediction requirements for different AV applications. Finally, the main limitations of our methodology and results are highlighted.

\subsection{Vehicle-level Performance Metrics}
\subsubsection{Safety} An overview of the pedestrian speeds and trajectory prediction horizons that resulted in a collision is depicted in Fig.~\ref{fig:results-safety}. In the figure, a cross indicates a simulation where the AV collided with the pedestrian, and the color the cross denotes the speed of the AV at the moment of collision. As the prediction horizon increases, the number of collisions and the AV collision speed rapidly decrease.
\textit{SC1} requires a minimum horizon of 0.8 seconds to avoid all collisions\footnote{Note that with a horizon of 0.6 seconds, some of the collisions were caused by the pedestrian not reacting to the vehicle, i.e. the vehicle stopped in time but the pedestrian collided on the side. This is a limitation of the simulations, where the pedestrian does not react to the vehicle in these cases.}. For \textit{SC2} and \textit{SC3}, all collisions are avoided with at least 1.2 and 1.6 seconds, respectively. 

\subsubsection{Comfort}
An overview of the braking comfort for each scenario is shown in Fig.~\ref{fig:results-comfort}. The figure shows the percentage of the total braking time that was comfortable, uncomfortable, or highly uncomfortable, as a function of the prediction horizon.

\paragraph{SC1}{Highly uncomfortable braking is minimized with a prediction horizon of 7 seconds, occurring less than 0.13\% of the time. Already with horizons between 5-9 seconds, this high discomfort stays below 0.8\%, increasing slightly for higher horizons due to the decreased update rate of the planner, but it remains under 3.4\% up to 15 seconds. 
With a horizon of 10 seconds, the share of uncomfortable braking is minimized (5\%), and the share of comfortable braking is maximized (93.6\%). Thus, for SC1 the recommended minimum horizon is 7 seconds. Horizons beyond 10 seconds do not present any additional comfort benefit.

\paragraph{SC2}{Highly uncomfortable braking is minimized with a horizon of 7 seconds, occurring less than 1.7\% of the time. With 6 seconds, highly uncomfortable braking is already under 1.93\%. Beyond 7 seconds it begins to increase again, but it remains under 5\% up to 12 seconds. With a horizon of 15 seconds, uncomfortable braking is minimized (10.5\%) and comfortable braking is maximized (84.2\%). Thus, for SC2 the recommended horizons are between 7 and 15 seconds.

\paragraph{SC3}{Highly uncomfortable braking is minimized with a prediction horizon of 10 seconds, occurring about 3.6\% of the time. Even with 9 seconds, highly uncomfortable braking is already is under 3.8\%. Beyond 10 seconds it begins to increase again, but it remains under 5.9\% up to 15 seconds. The share of uncomfortable braking that does not involve highly uncomfortable braking is minimized (13\%) with a horizon of 15 seconds, and at the same horizon the comfortable braking is maximized (81\%).  For optimal comfort in SC3, the optimal horizons are 10-15 seconds.

\subsubsection{Efficiency}
Fig.~\ref{fig:results-travel-time} presents the average increase in travel time, for all SCs and pedestrian speeds: P1, P2, P3, and an overall average. Generally, the trends are consistent across these scenarios. With slow pedestrian speeds (P1), there is a notable increase in travel delay. The maximum delay ranges from 27.3\% to 39.6\%, while the minimum delay varies between 19.1\% and 22\%, depending on the specific scenario. For P2 speeds, the maximum delay is between 18\% and 20.5\%, and the minimum delay lies between 7.4\% and 7.7\%. P3 speeds result in the least delay, with maximum values ranging from 4.5\% to 7.6\% and minimum values from 2.8\% to 3.1\%.

Overall, the maximum and minimum travel delays are observed with horizons ranging between 1.8 and 2 seconds and between 7 and 8 seconds, respectively, as detailed in Table~\ref{tab:efficiency-results}.
\begin{table}[htb]
\centering
\caption{Maximum and minimum mean travel time delay and respective prediction horizons}
\label{tab:efficiency-results}
\begin{tabular}{@{}l|cc|cc@{}}
\toprule
 & \multicolumn{2}{c|}{Maximum} & \multicolumn{2}{c}{Minimum} \\
 & $\Delta t$ [\%] & horizon [s] & $\Delta t$ [\%] & horizon [s] \\ \midrule
SC1 & 10.6 & 2 & 7.5 & 8 \\
SC2 & 15 & 1.8 & 9.4 & 7 \\
SC3 & 17.5 & 1.8 & 9.8 & 8 \\ \bottomrule
\end{tabular}
\end{table}
\begin{figure}[htb]
    \centering
    \includegraphics[width=\linewidth]{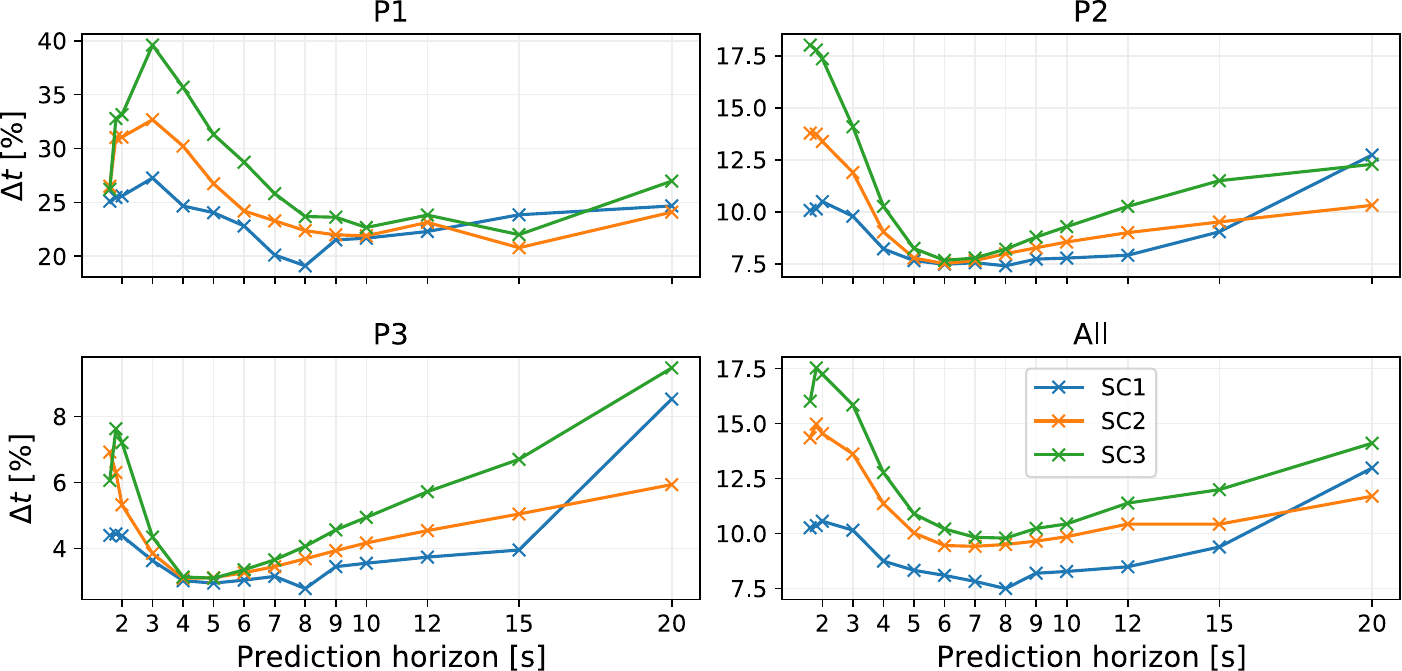}
    \caption{Mean travel time increase with different horizons for the three pedestrian speed groups, and overall. 
    }
    \label{fig:results-travel-time} 
    \vspace{-15pt}
\end{figure}

\subsection{Real-time Execution}
The mean and standard deviation of the minimum frequencies achieved by the trajectory planner for the different horizons are shown in Fig.~\ref{fig:results-frequency}. The mean of the minimum frequencies across runs is consistently kept above 10Hz up to and including 10 seconds of horizon for all three SCs. Due to computational overhead caused by planning for longer horizons, the minimum frequencies drop significantly beyond horizons 7, 6 and 5 seconds for SC1, SC2 and SC3 respectively. This drop in the throughput of trajectory generation has a negative impact on the stability of the AV, and consequently vehicle-level metrics, as shown next analysing the AV's acceleration during one of the simulations with various horizons.
\begin{figure}[htb]
    \centering
    \includegraphics[width=0.93\linewidth]{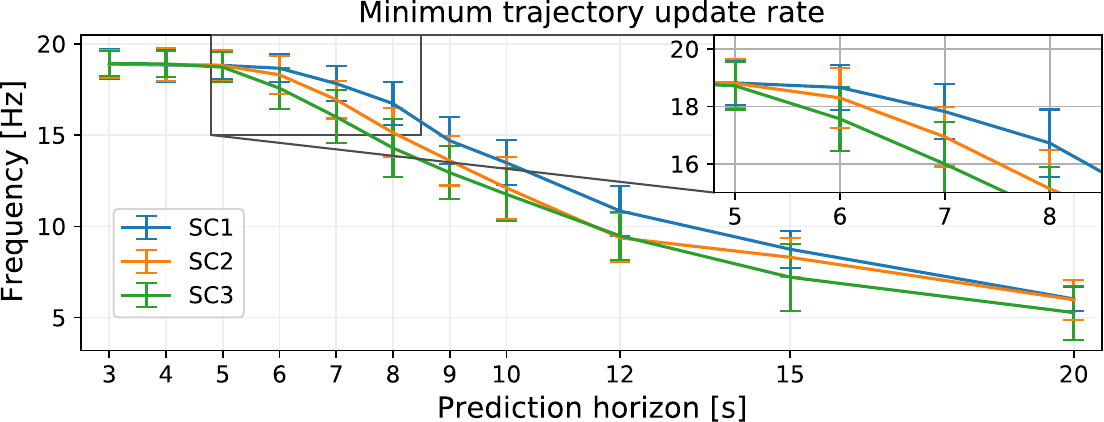}
    \caption{Update rate of the trajectory planner for each prediction horizon.}
    \label{fig:results-frequency}
\end{figure}

\noindent \textit{Comfort with increased computational load.} {Fig.~\ref{fig:frequency-example} shows an example of the impact that the computational load caused by longer horizons can have on comfort. With 10 seconds of horizon, the planner cannot maintain the configured 20Hz output, dropping somewhere between 12-20Hz when re-planning due to the pedestrian. This drop can cause some minor acceleration peaks, which are still maintained within the comfortable range. However, with a horizon of 20 seconds, the planning frequency drops below 10Hz, which causes significant oscillations in the AV's acceleration and result in uncomfortable braking.}
\begin{figure}[htb]
    \centering
    \includegraphics[width=0.93\linewidth]{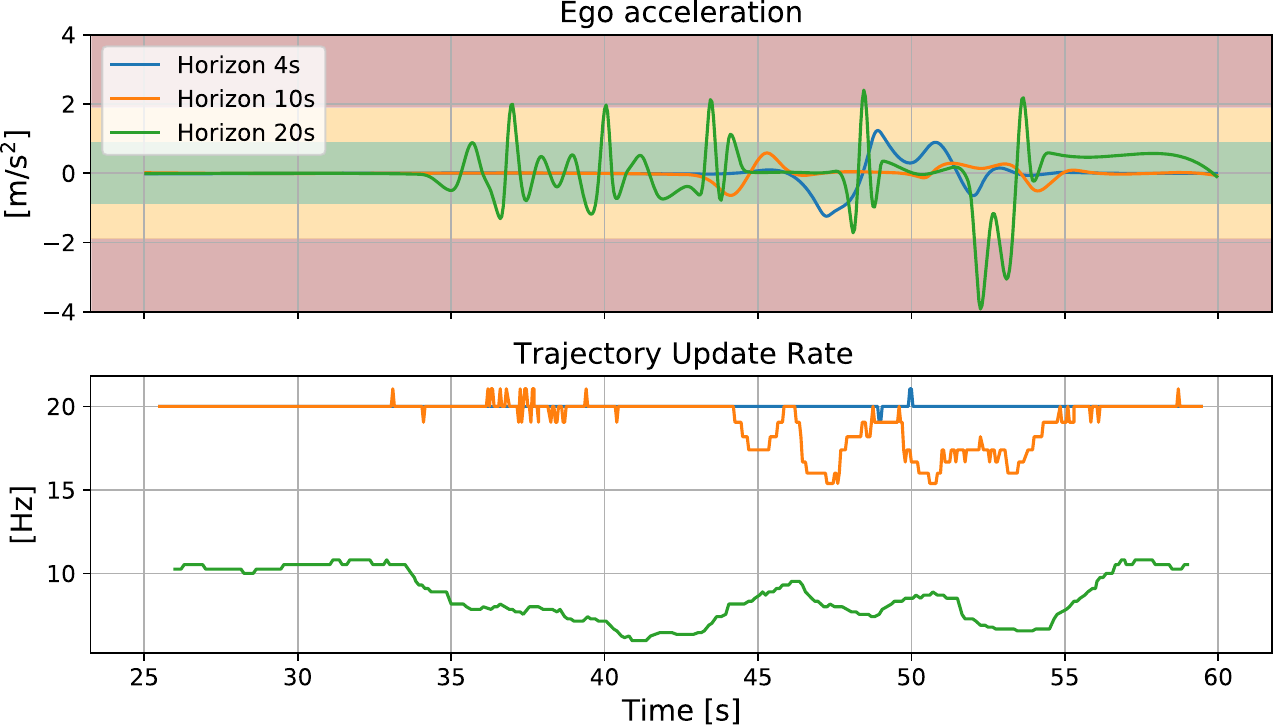}
    \caption{Example of the impact of increased computational load on the resulting acceleration due longer planning horizons. The green, yellow and red background on the top plot indicate the thresholds for comfortable, uncomfortable, and highly uncomfortable }
    \label{fig:frequency-example}
\end{figure}

\subsection{Application-dependent Prediction Horizons} 
The required and optimal prediction horizons for each SC and metric combination are summarized in Table~\ref{tab:results-overview}, including the overall horizons considering every SC and metric equally important. Note that some entries are empty, meaning that a required horizon satisfying all metrics in all scenarios could not be found.
Recall the four AV applications introduced earlier: \textit{a)} general-purpose urban driving, \textit{b)} driverless food delivery, \textit{c)} driverless taxi where the passenger is not in a hurry, and \textit{d)} driverless taxi where the passenger is in a hurry. Depending on the specific application, the required and optimal horizons vary significantly. According to the applications considered and the possible weighting scheme introduced in Table~\ref{tab:results-weights}, the required and optimal horizons achieved are shown in Table~\ref{tab:results-recommendations}. 
\begin{table}[htb]
\centering
\caption{Required and optimal horizons for each SC and metric. Calculated with \eqref{eq:aggregation:cost}-\eqref{eq:req}}
\label{tab:results-overview}
\begin{tabular}{@{}l|cc|cc|cc|cc@{}}
\toprule
 & \multicolumn{2}{|c|}{Safety} & \multicolumn{2}{c|}{Comfort} & \multicolumn{2}{c|}{Efficiency} & \multicolumn{2}{c}{Overall} \\ 
        & $\required_M^S$ & $\optimal_M^S$ & $\required_M^S$ & $\optimal_M^S$ & $\required_M^S$ & $\optimal_M^S$ & $\required_M^S$ & $\optimal_M^S$ \\ \midrule
SC1     & 0.8  & 0.8   & 7    & 10    & 4.1    & 8  & 7  & 8.8  \\
SC2     & 1.2  & 1.2   & 7    & 15    & 4.4    & 7  & 7 & 11.6  \\
SC3     & 1.6  & 1.6   & 10    & 15  & 4.3    & 8   & -  & 12  \\
Overall & 1.6  & 1.6   & 10    & 14.8   & 4.4    & 7.7 & -  & 11.8 \\ \bottomrule
\end{tabular}%
\end{table}
\begin{table}[htb]
\centering
\caption{Required and optimal horizons for each application. Calculated with \eqref{eq:aggregation:cost}-\eqref{eq:req}
}
\label{tab:results-recommendations}
\begin{tabular}{@{}l|cc@{}}
\toprule
\multicolumn{1}{c|}{Application} & $\required_M^S$ & $\optimal_M^S$ \\ \midrule
a) General-purpose urban driving & - & 11.8 \\
b) Driverless food delivery &  4.4 & 7.9 \\
c) Driverless taxi (not hurried) & 10 & 14.8 \\
d) Driverless taxi (hurried) & - & 8.9 \\ \bottomrule
\end{tabular}
\end{table}
Among the considered applications, driverless food delivery presents the shortest horizon requirements, with a required horizon of 4.4 seconds and an optimal horizon of 7.9 seconds. Driverless taxi applications would benefit from a minimum horizon of 10 seconds, and optimally 14.8 seconds if the passenger is not in a hurry. If the passenger is in a hurry, no horizon exists satisfying all requirements simultaneously, but optimally a horizon of 8.9 seconds would be selected. Finally, for a general-purpose AV operating in environments with crossing pedestrians, the optimal prediction horizon is 11.8 seconds.

\subsection{Limitations}\label{subsec:limitations}
The methodology presented in this work, and consequently our results regarding the recommended horizons for prediction models, has three main limitations.

Firstly, our work operates under the assumption of perfect predictions, which is highly unlikely in real-world driving. If prediction accuracy 
degrades rapidly with increased horizon, the resulting required and optimal horizons will be affected. It is challenging to discern the effect of these inaccuracies without further experiments, as they will depend on the considered vehicle-level metrics. For instance, inaccurate predictions might lead to more aggressive AV behavior, which would negatively affect comfort, while more aggressive acceleration could be beneficial for travel time efficiency. 

Secondly, the current study is constrained to scenarios involving crossing pedestrians and the use of a single planner. Given that different scenarios lead to distinct conclusions, it is vital to explore different scenarios and interactions with other RUs. Furthermore, our conclusions are tied to the specific trajectory planner used during this work. Integrating different planners into our study will enhance the generalizability of our conclusions. 

Lastly, the potential benefits of very long prediction horizons remain ambiguous due to computational constraints. It remains unclear whether the optimal value of each metric was achieved, and whether this optimal value would remain constant or degrade with longer horizons.

\section{Conclusion}\label{sec:conclusion}
Predicting the future movements of surrounding road users is essential for enhancing the performance of an automated vehicle (AV). However, the degree to which these predictions influence the AV's behavior is unknown.

In this study, we explore how various prediction horizons impact the behavior of an AV in terms of safety, comfort, and efficiency. We simulate trajectory predictions in crossing pedestrian scenarios and test different horizons of to 20 seconds, which are integrated into a state-of-the-art trajectory planner that considers the future motion of other road users. 

Our results indicate that a prediction horizon of 1.6 seconds is sufficient for avoiding collisions with crossing pedestrians in urban settings. Optimal travel time efficiency is achieved with longer horizons of 7-8 seconds, while predicting up to 15 seconds improves comfort. Longer horizons incur a higher computational load which negatively affects AV performance.

In this work we show that selecting a single optimal prediction horizon is not possible, as the best choice depends on the desired balance between different metrics and scenarios. To this end, we offer a framework to determine the required and optimal prediction horizons based on the specific AV application. Our study suggests a prediction horizon of 11.8 seconds as a general recommendation for AVs operating in environments with crossing pedestrians.


Future work will focus on extending our methodology to derive accuracy requirements for trajectory prediction models.

\printbibliography 

@article{Bae2019,
   author = {Il Bae and Jaeyoung Moon and Jeongseok Seo},
   doi = {10.3390/electronics8090943},
   issn = {2079-9292},
   issue = {9},
   journal = {Electronics},
   month = {8},
   pages = {943},
   title = {Toward a Comfortable Driving Experience for a Self-Driving Shuttle Bus},
   volume = {8},
   url = {https://www.mdpi.com/2079-9292/8/9/943},
   year = {2019},
}

@article{nhtsa2018critical,
  title={Critical Reasons for Crashes Investigated in the
National Motor Vehicle Crash Causation Survey},
  author={NHTSA},
  journal={{U.S. Department of Transportation}},
  year={2018},
  url={https://crashstats.nhtsa.dot.gov/Api/Public/ViewPublication/812506}
}

@inproceedings{D2.6,
 author = {B{\'a}lint, A and Labenski, V, and K{\"o}be, M. and Vogl, C. and Stoll, J. and  Schories, L. and  Amann, L. and  Sudhakaran, G. and Leyva, P. and Pallacci, T. and {\"O}stling, M. and Schmidt, D. and Schindler, R.},
 date = {11.10.2021},
 title = {{SAFE-UP} Deliverable Report: D2.6 USE CASE DEFINITIONS AND INITIAL SAFETY-CRITICAL SCENARIOS},
 url = {https://www.safe-up.eu/resources},
 urldate = {11.12.2023}
}

@techreport{pedestrianSpeeds,
	copyright = {In Copyright - Non-Commercial Use Permitted},
	year = {2006},
	volume = {132},
	type = {Report},
	journal = {IVT Schriftenreihe},
	author = {Buchmüller, Stefan and Weidmann, Ulrich},
	size = {56 p.},
	address = {Zürich},
	publisher = {Institut für Verkehrsplanung und Transportsysteme (IVT), ETH Zürich},
	DOI = {10.3929/ethz-b-000047950},
	title = {Parameters of pedestrians, pedestrian traffic and walking facilities}
}

@article{iso2631,
  title={{ISO 2631. Mechanical Vibration and Shock – Evaluation of Human Exposure to Whole-Body Vibration}},
  author={{ISO 2631:2001}},
  journal={International Organization for Standardization, Geneva},
  year={2001}
}

@inproceedings{MunozSanchez2022,
    title = {{Scenario-based Evaluation of Prediction Models for Automated Vehicles}},
    year = {2022},
    booktitle = {International Conference on Intelligent Transportation Systems},
    author = {Mu{\~{n}}oz S{\'{a}}nchez, Manuel and Elfring, Jos and Silvas, Emilia and van de Molengraft, Rene},
    pages = {2227--2233},
    url = {https://ieeexplore.ieee.org/document/9922603/},
    isbn = {978-1-6654-6880-0},
    doi = {10.1109/ITSC55140.2022.9922603},
    arxivId = {2210.06553}
}

@article{Schuetz2023AUser,
    title = {{A Review of Trajectory Prediction Methods for the Vulnerable Road User}},
    year = {2023},
    journal = {Robotics},
    author = {Schuetz, Erik and Flohr, Fabian B.},
    number = {1},
    month = {12},
    pages = {1},
    volume = {13},
    doi = {10.3390/robotics13010001},
    issn = {2218-6581}
}

@article{Taiebat2018AVehicles,
    title = {{A review on energy, environmental, and sustainability implications of connected and automated vehicles}},
    year = {2018},
    journal = {Environmental Science and Technology},
    author = {Taiebat, Morteza and Brown, Austin L. and Safford, Hannah R. and Qu, Shen and Xu, Ming},
    number = {20},
    pages = {11449--11465},
    volume = {52},
    doi = {10.1021/acs.est.8b00127},
    issn = {15205851},
    pmid = {30192527}
}

@article{GeisslingerAnVehicles,
  title={An ethical trajectory planning algorithm for autonomous vehicles},
  author={Geisslinger, Maximilian and Poszler, Franziska and Lienkamp, Markus},
  journal={Nature Machine Intelligence},
  volume={5},
  number={2},
  pages={137--144},
  year={2023},
  publisher={Nature Publishing Group UK London}
}

@article{Szimba2020AssessingRelation,
    title = {{Assessing travel time savings and user benefits of automated driving – A case study for a commuting relation}},
    year = {2020},
    journal = {Transport Policy},
    author = {Szimba, Eckhard and Hartmann, Martin},
    month = {11},
    pages = {229--237},
    volume = {98},
    doi = {10.1016/j.tranpol.2020.03.007},
    issn = {0967070X}
}

@article{Messaoud2021AttentionPrediction,
    title = {{Attention Based Vehicle Trajectory Prediction}},
    year = {2021},
    journal = {IEEE Transactions on Intelligent Vehicles},
    author = {Messaoud, Kaouther and Yahiaoui, Itheri and Verroust-Blondet, Anne and Nashashibi, Fawzi},
    number = {1},
    month = {3},
    pages = {175--185},
    volume = {6},
    doi = {10.1109/TIV.2020.2991952},
    issn = {2379-8904}
}

@article{Ercan2022AutonomousCities,
    title = {{Autonomous electric vehicles can reduce carbon emissions and air pollution in cities}},
    year = {2022},
    journal = {Transportation Research Part D: Transport and Environment},
    author = {Ercan, Tolga and Onat, Nuri C. and Keya, Nowreen and Tatari, Omer and Eluru, Naveen and Kucukvar, Murat},
    month = {11},
    pages = {103472},
    volume = {112},
    doi = {10.1016/j.trd.2022.103472},
    issn = {13619209}
}

@article{Dixit2016AutonomousTimes,
    title = {{Autonomous Vehicles: Disengagements, Accidents and Reaction Times}},
    year = {2016},
    journal = {PLOS ONE},
    author = {Dixit, Vinayak V. and Chand, Sai and Nair, Divya J.},
    number = {12},
    month = {12},
    volume = {11},
    doi = {10.1371/journal.pone.0168054},
    issn = {1932-6203}
}

@inproceedings{Li2020AV-FUZZER:Systems,
    title = {{AV-FUZZER: Finding Safety Violations in Autonomous Driving Systems}},
    year = {2020},
    booktitle = {IEEE ISSRE},
    author = {Li, Guanpeng and Li, Yiran and Jha, Saurabh and Tsai, Timothy and Sullivan, Michael and Hari, Siva Kumar Sastry and Kalbarczyk, Zbigniew and Iyer, Ravishankar},
    month = {10},
    pages = {25--36},
    isbn = {978-1-7281-9870-5},
    doi = {10.1109/ISSRE5003.2020.00012}
}

@inproceedings{vanderPloeg2023ConnectingReasoning,
    title = {{Connecting the Dots: Context-Driven Motion Planning Using Symbolic Reasoning}},
    year = {2023},
    booktitle = {IEEE Intelligent Vehicles Symposium},
    author = {van der Ploeg, Chris and Braat, Michiel and Masini, Beatrice and Brouwer, Jochem and Paardekooper, Jan-Pieter},
    month = {6},
    pages = {1--8},
    isbn = {979-8-3503-4691-6},
    doi = {10.1109/IV55152.2023.10186794}
}

@article{Jungblut2023Fuel-savingAnalysis,
    title = {{Fuel-saving opportunities for automated vehicles: A driving cycle analysis}},
    year = {2023},
    journal = {Transportation Research Interdisciplinary Perspectives},
    author = {Jungblut, Edgar and Grube, Thomas and Linssen, Jochen and Stolten, Detlef},
    month = {11},
    pages = {100964},
    volume = {22},
    doi = {10.1016/j.trip.2023.100964},
    issn = {25901982}
}

@article{Wadud2023FullyUse,
    title = {{Fully automated vehicles: the use of travel time and its association with intention to use}},
    year = {2023},
    journal = {Proceedings of the Institution of Civil Engineers - Transport},
    author = {Wadud, Zia and Huda, Fuad Yasin},
    number = {3},
    month = {6},
    pages = {127--141},
    volume = {176},
    doi = {10.1680/jtran.18.00134},
    issn = {0965-092X}
}

@inproceedings{Dong2022Graph-basedDriving,
author = {Dong, Qing and Jiang, Titong and Xu, Tao and Liu, Yahui},   
month = {10},

   title = {{Graph-based Planning-informed Trajectory Prediction for Autonomous Driving}},
   year = {2022},
booktitle = {CAA CVCI},
}

@article{Rudenko2020HumanSurvey,
    title = {{Human motion trajectory prediction: a survey}},
    year = {2020},
    journal = {The International Journal of Robotics Research},
    author = {Rudenko, Andrey and Palmieri, Luigi and Herman, Michael and Kitani, Kris M. and Gavrila, Dariu M. and Arras, Kai O.},
    number = {8},
    month = {7},
    pages = {895--935},
    volume = {39},
    doi = {10.1177/0278364920917446},
    issn = {0278-3649},
    arxivId = {1905.06113}
}

@inproceedings{Ivanovic2022InjectingEvaluation,
    title = {{Injecting Planning-Awareness into Prediction and Detection Evaluation}},
    year = {2022},
    booktitle = {IEEE Intelligent Vehicles Symposium},
    author = {Ivanovic, Boris and Pavone, Marco},
    month = {6},
    pages = {821--828},
    url = {https://ieeexplore.ieee.org/document/9827101/},
    isbn = {978-1-6654-8821-1},
    doi = {10.1109/IV51971.2022.9827101}
}

@InProceedings{IvanovicMATS:Control,
  title = 	 {{MATS: An Interpretable Trajectory Forecasting Representation for Planning and Control}},
  author =       {Ivanovic, Boris and Elhafsi, Amine and Rosman, Guy and Gaidon, Adrien and Pavone, Marco},
  booktitle = 	 {{CoRL}},
  pages = 	 {2243--2256},
  year = 	 {2021},
  
}

@inproceedings{Messaoud2019Non-localPrediction,
    title = {{Non-local Social Pooling for Vehicle Trajectory Prediction}},
    year = {2019},
    booktitle = {IEEE Intelligent Vehicles Symposium},
    author = {Messaoud, Kaouther and Yahiaoui, Itheri and Verroust-Blondet, Anne and Nashashibi, Fawzi},
    month = {6},
    pages = {975--980},
    isbn = {978-1-7281-0560-4},
    doi = {10.1109/IVS.2019.8813829}
}

@article{Sohrabi2021QuantifyingResearch,
    title = {{Quantifying the automated vehicle safety performance: A scoping review of the literature, evaluation of methods, and directions for future research}},
    year = {2021},
    pages={106003},
    journal = {Accident Analysis and Prevention},
    author = {Sohrabi, Soheil and Khodadadi, Ali and Mousavi, Seyedeh Maryam and Dadashova, Bahar and Lord, Dominique},
    volume = {152},
    publisher = {Elsevier Ltd},
    url = {https://doi.org/10.1016/j.aap.2021.106003},
    doi = {10.1016/j.aap.2021.106003},
    issn = {00014575},
    pmid = {33571922}
}

@inproceedings{Hagedorn2023RethinkingReview,
      title={{Rethinking Integration of Prediction and Planning in Deep Learning-Based Automated Driving Systems: A Review}}, 
      author={Steffen Hagedorn and Marcel Hallgarten and Martin Stoll and Alexandru Condurache},
      year={2023},
      eprint={2308.05731},
      archivePrefix={arXiv},
}

@inproceedings{Gupta2018SocialNetworks,
    title = {{Social GAN: Socially Acceptable Trajectories with Generative Adversarial Networks}},
    year = {2018},
    booktitle = {IEEE Computer Vision and Pattern Recognition},
    author = {Gupta, Agrim and Johnson, Justin and Fei-Fei, Li and Savarese, Silvio and Alahi, Alexandre},
    month = {6},
    pages = {2255--2264},
    isbn = {978-1-5386-6420-9},
    doi = {10.1109/CVPR.2018.00240},
    issn = {10636919},
    arxivId = {1803.10892}
}

@inproceedings{Alahi2016SocialSpaces,
   author = {Alexandre Alahi and Kratarth Goel and Vignesh Ramanathan and Alexandre Robicquet and Li Fei-Fei and Silvio Savarese},
   doi = {10.1109/CVPR.2016.110},
   isbn = {978-1-4673-8851-1},
   issn = {10636919},
   booktitle = {IEEE CVPR},
   month = {6},
   pages = {961-971},
   title = {{Social LSTM: Human Trajectory Prediction in Crowded Spaces}},
   url = {http://ieeexplore.ieee.org/document/7780479/},
   year = {2016},
}

@inproceedings{Sadeghian2019SoPhie:Constraints,
   author = {Amir Sadeghian and Vineet Kosaraju and Ali Sadeghian and Noriaki Hirose and Hamid Rezatofighi and Silvio Savarese},
   doi = {10.1109/CVPR.2019.00144},
   isbn = {978-1-7281-3293-8},
   issn = {10636919},
   booktitle = {IEEE Computer Vision and Pattern Recognition},
   month = {6},
   pages = {1349-1358},
   title = {{SoPhie: An Attentive GAN for Predicting Paths Compliant to Social and Physical Constraints}},
   url = {https://ieeexplore.ieee.org/document/8953374/},
   year = {2019},
}

@inproceedings{Zhang2019SR-LSTM:Prediction,
   author = {Pu Zhang and Wanli Ouyang and Pengfei Zhang and Jianru Xue and Nanning Zheng},
   doi = {10.1109/CVPR.2019.01236},
   isbn = {978-1-7281-3293-8},
   issn = {10636919},
   booktitle = {IEEE Computer Vision and Pattern Recognition},
   month = {6},
   pages = {12077-12086},
   title = {{SR-LSTM: State Refinement for LSTM Towards Pedestrian Trajectory Prediction}},
   url = {https://ieeexplore.ieee.org/document/8954402/},
   year = {2019},
}

@article{deWinkel2023StandardsJerk,
    title = {{Standards for passenger comfort in automated vehicles: Acceleration and jerk}},
    year = {2023},
    journal = {Applied Ergonomics},
    author = {de Winkel, Ksander N. and Irmak, Tugrul and Happee, Riender and Shyrokau, Barys},
    month = {1},
    volume = {106},
    publisher = {Elsevier Ltd},
    doi = {10.1016/j.apergo.2022.103881},
    issn = {00036870},
    pmid = {36058166}
}

@inproceedings{Moers2022TheGermany,
    title = {{The exiD Dataset: A Real-World Trajectory Dataset of Highly Interactive Highway Scenarios in Germany}},
    year = {2022},
    booktitle = {IEEE Intelligent Vehicles Symposium,},
    author = {Moers, Tobias and Vater, Lennart and Krajewski, Robert and Bock, Julian and Zlocki, Adrian and Eckstein, Lutz},
    pages = {958--964},
    isbn = {9781665488211},
    doi = {10.1109/IV51971.2022.9827305}
}

@article{Petrovic2020TrafficDrivers,
    title = {{Traffic Accidents with Autonomous Vehicles: Type of Collisions, Manoeuvres and Errors of Conventional Vehicles’ Drivers}},
    year = {2020},
    journal = {Transportation Research Procedia},
    author = {Petrovi{\'{c}}, Dorđe and Mijailovi{\'{c}}, Radomir and Pe{\v{s}}i{\'{c}}, Dalibor},
    pages = {161--168},
    volume = {45},
    doi = {10.1016/j.trpro.2020.03.003},
    issn = {23521465}
}

@inproceedings{Ivanovic2023Trajdata:Datasets,
      title={{trajdata: A Unified Interface to Multiple Human Trajectory Datasets}}, 
      author={Boris Ivanovic and Guanyu Song and Igor Gilitschenski and Marco Pavone},
      year={2023},
      eprint={2307.13924},
      archivePrefix={arXiv},
}

@inproceedings{Mozaffari2023TrajectoryScenarios,
      title={{Trajectory Prediction with Observations of Variable-Length for Motion Planning in Highway Merging scenarios}}, 
      author={Sajjad Mozaffari and Mreza Alipour Sormoli and Konstantinos Koufos and Graham Lee and Mehrdad Dianati},
      year={2023},
      eprint={2306.05478},
      archivePrefix={arXiv},
}

@inproceedings{Tran2023WhatDriving,
 author = {Phong, Tran and Wu, Haoran and Yu, Cunjun and Cai, Panpan and Zheng, Sifa and Hsu, David},
 booktitle = {{NeurIPS}},
 pages = {71327--71339},
 title = {{What Truly Matters in Trajectory Prediction for Autonomous Driving?}},
 year = {2023}
}

@article{Artinger2022Satisficing:Traditions,
    title = {{Satisficing: Integrating Two Traditions}},
    year = {2022},
    journal = {Journal of Economic Literature},
    author = {Artinger, Florian M. and Gigerenzer, Gerd and Jacobs, Perke},
    number = {2},
    month = {6},
    pages = {598--635},
    volume = {60},
    doi = {10.1257/jel.20201396},
    issn = {0022-0515}
}


\end{document}